\documentclass[final,1p,times,authoryear]{elsarticle}
\usepackage{amssymb}
\usepackage{amsmath}
\usepackage{url}
\usepackage[hidelinks, hyperfootnotes=false]{hyperref}

\usepackage{multirow}    % For multirow cells
\usepackage{array}       % For better table formatting
\usepackage{booktabs}    % For elegant horizontal lines (midrule)
\usepackage{siunitx}     % For better alignment of numbers
\usepackage{float}

\makeatletter
\def\ps@pprintTitle{%
    \let\@oddhead\@empty
    \let\@evenhead\@empty
    \let\@oddfoot\@empty
    \let\@evenfoot\@empty
}
\makeatother

\begin{document}

\begin{frontmatter}


\title{BDA: Bangla Text Data Augmentation Framework} %% Article title

%% Author information with specific emails

%% Authors
\author[ssl]{Md. Tariquzzaman\corref{cor1}}
\ead{tariquzzaman@iut-dhaka.edu}

\author[ssl]{Audwit Nafi Anam}
\ead{audwitnafi@iut-dhaka.edu}

\author[ssl]{Naimul Haque}
\ead{naimulhaque@iut-dhaka.edu}

\author[ssl,manchester]{Mohsinul Kabir}
\ead{mohsinulkabir@iut-dhaka.edu}

\author[ssl]{Hasan Mahmud}
\ead{hasan@iut-dhaka.edu}

\author[ssl]{Md Kamrul Hasan}
\ead{hasank@iut-dhaka.edu}

%% Affiliations
\affiliation[ssl]{organization={Systems and Software Lab (SSL), Department of Computer Science and Engineering, Islamic University of Technology},
            addressline={K B Bazar Road}, 
            city={Gazipur}, 
            postcode={1704}, 
            country={Bangladesh}}

\affiliation[manchester]{organization={The University of Manchester}, 
                         addressline={Oxford Road}, 
                         city={Manchester}, 
                         postcode={M13 9PL}, 
                         country={United Kingdom}}

%% Corresponding author footnote
\cortext[cor1]{Corresponding author: Md. Tariquzzaman, email: tariquzzaman@iut-dhaka.edu}

%% Abstract
\begin{abstract}
Data augmentation involves generating synthetic samples that resemble those in a given dataset. In resource-limited fields where high-quality data is scarce, augmentation plays a crucial role in increasing the volume of training data. This paper introduces a Bangla Text Data Augmentation (BDA) Framework that uses both pre-trained models and rule-based methods to create new variants of the text. A filtering process is included to ensure that the new text keeps the same meaning as the original while also adding variety in the words used. We conduct a comprehensive evaluation of the framework's effectiveness in Bangla text classification tasks. Our framework achieved significant improvement in F1 scores across five distinct datasets, delivering performance equivalent to models trained on 100\% of the data while utilizing only 50\% of the training dataset. Additionally, we explore the impact of data scarcity by progressively reducing the training data and augmenting it through BDA, resulting in notable F1 score enhancements. The study offers a thorough examination of BDA's performance, identifying key factors for optimal results and addressing its limitations through detailed analysis.
\end{abstract}

% % \includegraphics{grabs}
% \begin{figure}[h]
%     \centering
%     \includegraphics[width=\textwidth]{BDA_Bangla Text Data Augmentation Framework_for_arxiv/figs/Graphical_Abstract.pdf} % Insert the first page of the PDF
%     \caption{Graphical Abstract}
% \end{figure}

% % %% Research highlights
% \begin{highlights}
% \item Development of BDA, a comprehensive Bangla text data augmentation framework combining rule-based and transformer-based methods for generating semantically equivalent augmented samples.
% \item Introduction of a filtering mechanism to ensure high-quality, meaningful augmentation while preserving the original text's intent.
% \item Detailed evaluation of BDA's effectiveness in Bangla text classification, demonstrating significant improvements in F1 scores with only 50\% of the original data.
% \item In-depth analysis of the impact of data scarcity on Bangla NLP tasks and how BDA addresses this challenge through augmentation.
% \item Identification of optimal contexts for Bangla text augmentation, and discussion of scenarios where it might be less effective.
% \end{highlights}

%% Keywords
% Keywords section
\end{frontmatter}



\section{Introduction}

Bangla, the seventh most spoken language\footnote{\url{https://en.wikipedia.org/wiki/List_of_languages_by_total_number_of_speakers}} globally, requires increasingly sophisticated tools for effective understanding and processing. However, there is a lack of robust augmentation frameworks for Bangla that maintain diversity while preserving the original meaning. Existing datasets, as explored by \citet{banglabook}, are mostly small and exhibit low lexical variance among samples. While developing a large, high-quality dataset would be optimal, it is often impractical due to the extensive annotation required. Therefore, an efficient alternative is needed. Text data augmentation offers an effective solution, having shown potential to improve the performance of Natural Language Processing models. It artificially expands training datasets by generating additional examples that are coherent with the original data and capture the nuances of the language. This approach is particularly crucial for Bangla, as it enables us to overcome limitations imposed by the scarcity of Bangla text datasets and enhances the generalization capabilities and robustness of NLP models.

Currently, Bangla lacks a thoroughly evaluated text augmentation framework that systematically leverages multiple augmentation approaches to maximize the diversity and quality of augmented texts. Existing solutions rely on a limited set of techniques, highlighting a pressing need for a more effective framework. To address this, we introduce BDA\footnote{\url{https://github.com/tzf101/Bangla-Text-Augmentation-Framework}}, a Bangla text augmentation framework that incorporates a combination of model-based and rule-based approaches to generate a wide array of synthetic samples with semantically equivalent variations, along with a filtering mechanism to remove low-quality outputs. By filling this critical gap, we believe the BDA framework can significantly advance the field of Bangla NLP, bridging the data scarcity of high-quality datasets. Our main contributions through the BDA framework are as follows:
\begin{enumerate}
    \item We develop a comprehensive Bangla text augmentation pipeline that integrates both rule-based and transformer-based methods to generate a wide variety of augmented texts, coupled with a filtering process to ensure high-quality outputs.
    \item We provide an in-depth analysis of the effectiveness of augmentation in the Bangla language, highlighting the individual impact of each method and explaining the necessity of using a diverse set of techniques rather than relying on a single approach.
    \item We offer a detailed evaluation of the contexts in which Bangla text augmentation performs optimally and where it may be less effective.
\end{enumerate}
\newcounter{firstfootnote}
\setcounter{firstfootnote}{\value{footnote}}

\section{Related Work}

Data augmentation (DA) has been extensively studied and applied in computer vision, with techniques such as cropping, flipping, and rotation \citep{imageaug}. These methods have been highly effective in addressing issues like overfitting and improving model generalization. However, in Natural Language Processing (NLP), especially for languages like Bangla, DA techniques have only been explored more recently.

The application of data augmentation to text is more challenging due to the discrete and complex nature of natural language. Simple methods like word swapping, insertion, or synonym replacement can often lead to issues. For instance, not all words can be freely replaced, particularly function words (e.g., articles, prepositions, conjunctions) or common nouns, which may not have suitable synonyms. Furthermore, replacing words without careful consideration can alter the meaning or structure of a sentence, leading to incorrect or ambiguous interpretations. In such cases, the language model may fail to classify or understand the text properly. Although some research has been conducted on text augmentation for various languages, there is a notable lack of work on effective frameworks for Bangla text augmentation. This is particularly significant for resource-scarce languages like Bangla  \citep{mukherjee-etal-2023-low}, where augmenting available data is crucial for improving model performance. Despite the growing interest in NLP for Bangla, existing studies have not sufficiently addressed the specific challenges and needs for augmentation techniques tailored to this language.

The effectiveness of data augmentation relies on the underlying methods being used. While there is a noticeable lack of studies on Bangla-specific augmentation frameworks, several foundational building blocks for augmentation in Bangla are readily available. These resources have allowed us to implement BDA, an augmentation pipeline tailored for Bangla NLP tasks.

A simple way to solve skewed data or scarcity of data is through oversampling the existing dataset. But this leads to overfitting and has very little scope to improve the performance as demonstrated in \citet{oversamplingw2vreplace}. A better approach is to change parts of the text so that it still mimics the original sentence's inherent meaning. Such an appraoch is Synonym Replacement, a rule based approach where a word is replaced by its synonym to generate an augmented sentence. It's effectiveness has been studied in \citet{Sultana2024EnhancingBC} where they used it to increase the training dataset through augmentation, which was able to improve their benchmark scores in a Bangla text dataset.  

One of the widely used augmentation techniques is back-translation, which has been explored for Bangla by \citet{hasan-etal-2020-low}. They proposed a high-quality Bangla-English parallel corpus for Machine Translation (MT), created using a custom segmenter and novel alignment and filtering methods. Additionally, a pre-trained checkpoint of BanglaT5 \cite{bhattacharjee2023banglanlg}, a sequence-to-sequence Transformer model, has been fine-tuned on the BanglaNMT dataset for Bangla-English translation tasks, providing an essential resource for back-translation-based augmentation. This is proven to work well in practice as showed in \citet{riyad-etal-2023-team} where they used back-translation to augment the training dataset to mitigate scarcity of data in a Bangla dataset for a sentiment analysis task. They also discussed the rational of including emoji's and stopwords in Bangla since they may alter the meaning of the text due to the complex nature of the Bangla Language. Similar studies in Bangla NLP includes BanglaParaphrase by \citet{akil-etal-2022-banglaparaphrase} where they synthetically generated a high-quality corpus focused on lexical diversity and semantic similarity. A pre-trained BanglaT5 model, fine-tuned on this dataset, is also available, making it a valuable tool for paraphrasing-based augmentation. 

Augmented text can sometimes deviate from their intrinsic meaning due to changes in sentence structure and introduction of noise, which leads to shifting the class label of the augmented text, even though they should remain the same. This issue is discussed in \citet{wang-etal-2024-sta} where they mentioned about implementing a self-checking procedure, which emphasizes the fact that a proper augmentation pipeline must consist more than just augmented text generation, rather a proper check mechanism to ensure good quality augmentation, something that is not yet tested in case of Bangla text augmentation. \\

There is a growing need for developing Bangla text augmentation frameworks due to Bangla being a resource scarce language. In works such as \citet{das-etal-2023-team-error} we can see the potential of Bangla text augmentation. Existing augmentation toolkits such as \texttt{bnaug}\footnote{\url{https://github.com/sagorbrur/bnaug}} and \texttt{banglanlptoolkit}\footnote{\url{https://pypi.org/project/banglanlptoolkit}} offer standard functionalities such as back-translation, paraphrasing, token replacement via fill-mask, and random augmentation. However, they have limitations, including a lack of mechanisms to prevent label flipping in classification tasks and the absence of benchmarks to validate the effectiveness of the generated augmentations. Furthermore, token replacement methods are susceptible to label-changing issues, as discussed by \citet{iml}, and these toolkits often lack adequate filtering mechanisms to ensure that the augmented data introduces meaningful variation while preserving the original label and semantic integrity.

Previous studies, such as \citet{augvic}, have primarily focused on isolated techniques like back-translation without an in-depth analysis of their impact on Bangla NLP datasets. Our study addresses these gaps by systematically evaluating the combined impact of these existing tools and techniques on Bangla NLP tasks, ensuring that the augmented text enhances both quality and performance.

\section{BDA}
The BDA (Bangla text Data Augmentation) framework is designed to generate synthetic data that closely resembles the original dataset while preserving the underlying meaning of the texts. A dataset is passed through the BDA framework, where texts are augmented through these methods randomly and we get an augmented dataset, introducing a variety of new samples with different grammatical structures. It seeks to address the challenge of data scarcity in limited-resource datasets, where the small data size hinders the performance of classification models. To address this, we explored several augmentation techniques in BDA to expand the dataset and assess their effect on model performance. Since we are working with Bangla, we need models that are specific to this language. For each sentence in the training set, we performed the following operations:

\begin{enumerate}
    \item \textbf{Synonym Replacement (SR):} Randomly select \textit{n} non-stop words from the sentence and replace them with synonyms chosen from the most similar alternatives.
    \item \textbf{Random Swap (RS):} Randomly select two words from the sentence and swap their positions. This process is repeated \textit{n} times.
    \item \textbf{Back-Translation (BT):} Translate the Bangla text into an intermediate language (e.g., English) and then translate it back into Bangla.
    \item \textbf{Paraphrasing (PP):} Rephrase the original Bangla text while preserving its semantic meaning.
\end{enumerate}

For Synonym Replacement (SR), we utilized the Bangla Word2vec embeddings from the bnlp-toolkit\footnote{\url{https://github.com/sagorbrur/bnlp}} that searches for the nearest word vector to a target word and replace it as a synonym. For Back-Translation (BT), we employ the pre-trained BanglaT5 model checkpoint, fine-tuned on the BanglaNMT dataset \cite{hasan-etal-2020-low}, to generate translations. The process involves translating a Bangla text sample to English, then reverting it back to Bangla. During this process, the grammatical structure changes due to the nature of the languages. The Paraphrasing (PP) method follows a similar approach to BT. Instead, it uses a different alignment process, leveraging the paraphrase checkpoint of BanglaT5, which has been fine-tuned on the BanglaParaphrase dataset \cite{akil-etal-2022-banglaparaphrase}. This will allow us to address language-specific issues. 

% \begin{table}[h]
% \centering
%     \begin{tabular}{{{p{.2\columnwidth}c}p{.71\columnwidth}}}
% \hline
% \textbf{Operation} & \textbf{Sentence} \\
% \hline
% \textbf{Normal} & \textbengali{সঠিক তদন্ত করতে হবে বিচারের আওতায় আনতে হবে যে এই কাজ টা করেছ}. \\
% \textbf{SR} & \textbengali{সঠিক \textbf{আপিল} করতে হবে বিচারের আওতায় দিতে হবে যে এই কাজ টা করেছে} 
% \\
% \textbf{RS} & \textbengali{সঠিক তদন্ত করতে হবে বিচারের আওতায় \textbf{হবে আনতে} যে এই কাজ টা করেছে} 
% \\
% \textbf{BT} & \textbengali{সঠিক তদন্ত অবশ্যই করতে হবে এবং যে ব্যক্তি এটা করেছে তাকে বিচারের আওতায় আনতে হবে} 
% \\
% \textbf{PP} & \textbengali{সঠিক তদন্ত করতে হবে এবং এই কাজটি বিচারের আওতায় আনতে হবে যে এটি করেছে} 
% \\
% \hline

%     \end{tabular}
%     \caption{Sentences generated using BDA}
%     \label{tab:bda_examples}
% \end{table}

\begin{figure}[H]
    \centering
    \includegraphics[width=0.92\textwidth]{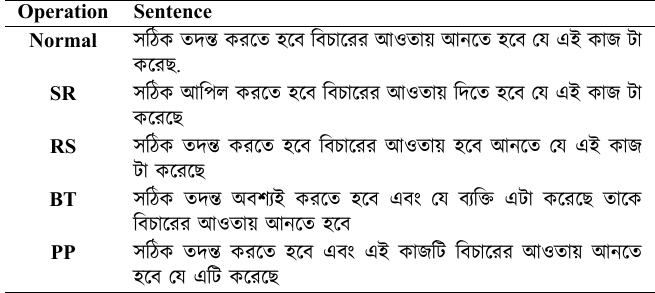}
    \caption{Sentences generated using BDA}
\end{figure}

Furthermore, rule-based models are particularly useful for Bangla due to the nature of Bangla informal social media texts where there is no particular grammatical structure. Another issue regarding Bangla is out of vocabulary words, written mostly in the wrong spelling. In these cases, model-based approaches fail to augment and thus rule-based approaches such as random swap can still work, as understanding the underlying meaning is not handled during augmentation, but rather handled by the filtering process.

To control the amount of augmentation, we process the entire dataset through BDA, creating a 1:1 augmented version by randomly selecting augmentation methods and merging it with the original dataset. As a result, the augmented dataset contains twice the number of samples, with half originating from BDA. We avoid further augmentation because it would result in over-representing synthetic data, which could skew the dataset and potentially lead to model bias towards augmented samples. This imbalance might degrade the model's ability to generalize effectively to real-world, unaltered data. For rule-based methods (Synonym Replacement and Random Swap), we select a small augmentation value (\textit{n}=2), as recommended by \cite{wei-zou-2019-eda}, to maintain the semantic integrity of the original sentences while introducing diversity. A smaller $n$ ensures that the augmented data does not deviate significantly from the original context, preventing potential noise that could adversely affect model training. This approach balances the need for data augmentation with the preservation of data quality, ultimately enhancing model performance without compromising accuracy.

\begin{figure}[h]
    \centering
    \includegraphics[width=10cm]{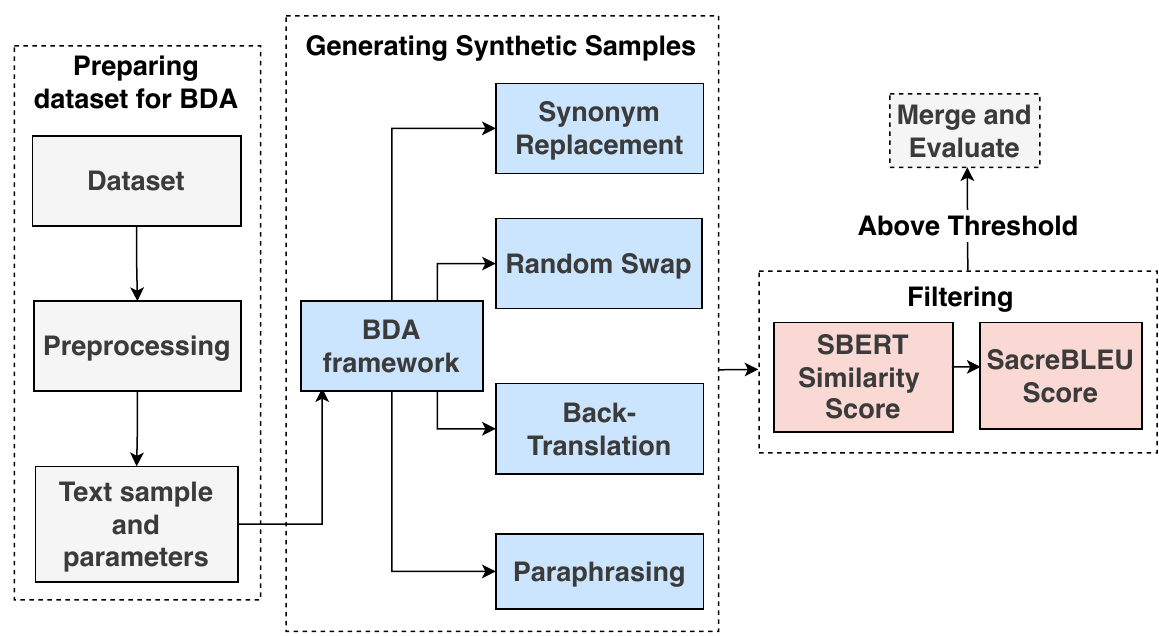}
    \caption{BDA Pipeline}
    \label{fig:bda-pipeline}
\end{figure}

\section{Experimental Setup}

To evaluate the effectiveness of the BDA framework, we designed a comprehensive test environment. We selected five datasets with diverse characteristics to ensure our results are thorough and applicable to various scenarios. These datasets represent a wide range of augmentation challenges, such as class imbalance, data scarcity, and low variation among samples. To examine the impact of data scarcity, we applied stratified sampling at 15\%, 50\%, and 100\% of the original dataset size. In stratified sampling, the population or dataset is divided into distinct ``strata" or subgroups based on class labels from where samples are drawn from each subgroup in a way that maintains the proportions of these subgroups in the overall dataset. This allows us to observe how augmentation affects performance across different levels of data availability in the BDA framework, without altering the original class ratio. The training dataset is clipped at 15\%, 50\%, and 100\% to evaluate how effectively data augmentation can replicate the performance achieved with the full dataset. This allows us to measure the extent to which augmentation can compensate for reduced data availability.

We set thresholds to filter out bad augmented texts that might be generated from BDA, depending on the dataset. Since these are dataset-dependent, in our experiments, we set the semantic similarity threshold between 85\% and 99\% to keep our results consistent. This range allows us to exclude texts that deviate too much from the original meaning, as well as those that are almost identical. Regarding lexical similarity, lower scores indicate greater diversity; therefore, we consider texts with a lexical similarity below 45\% to be acceptable. Since these values are specific to our dataset, they can be adjusted according to different requirements.

\subsection{Benchmark Datasets}
We selected five standard datasets to evaluate BDA, summarized in Table~\ref{tab:benchmark_datasets}. Each dataset was selected for its unique properties, providing a broad range of testing scenarios for the BDA framework. These datasets cover a broad spectrum of linguistic tasks and augmentation difficulties, allowing for a comprehensive assessment of the BDA framework’s performance across various real-world challenges.

\begin{table}[H]
\centering
\small
    \begin{tabular}
    {>{\centering\arraybackslash}m{.18\textwidth} m{.55\textwidth} >{\centering\arraybackslash}m{.12\textwidth} >{\centering\arraybackslash}m{.1\textwidth}}
\hline
\textbf{Dataset}          & \textbf{Description} & \textbf{Samples} & \textbf{Classes} \\ \hline
\textbf{BemoC} \cite{BEmoC} & A large dataset with six distinct labels, offering a robust multi-class classification problem. Its size and diversity make it ideal for testing BDA’s effectiveness in handling large-scale, linguistically varied data. & 48,328  & 6 \\ \hline
\textbf{SentNoB} \cite{sentnob} & Contains noisy, real-world text, simulating the challenge of working with informal, unstructured Bangla text. It is essential for evaluating BDA’s ability to handle noisy data effectively. & 15,728  & 3 \\ \hline
\textbf{Bengali Sentiment} \cite{benglai_sentiment} & Provides a balanced sentiment classification task, commonly used as a benchmark for sentiment analysis in Bangla. It serves as a baseline for evaluating BDA’s improvement in standard sentiment classification tasks. & 11,851  & 3 \\ \hline
\textbf{ABSA Dataset (Cricket)} \cite{ABSA} & Identifies sentiment polarity (positive, negative, neutral) and also analyzes specific aspects (e.g., ``service," ``food" in a review). Challenges the BDA framework in handling nuanced contextual dependencies. & 1,787   & 5 \\ \hline
\textbf{ABSA Dataset (Restaurant)} \cite{ABSA} & Similar to the cricket version, focuses on aspect-based sentiment analysis for restaurant reviews. Tests the ability of the framework to manage detailed sentiment aspect relationships. & 1,235   & 5 \\ \hline
\end{tabular}
\caption{Benchmark Datasets with Descriptions}
\label{tab:benchmark_datasets}
\end{table}
\subsection{Text Classification Models}

To evaluate the effectiveness of BDA in text classification tasks, we employed two distinct models: BanglaBERT \cite{BanglaBERT} and a Support Vector Machine (SVM). The SVM model was tested using word $n$-grams (1-3) and character $n$-grams (2-5). Each instance was converted into feature vectors through TF-IDF (Term Frequency-Inverse Document Frequency) weighting applied to the respective $n$-grams. For BanglaBERT, we kept the following parameters as shown in Table \ref{tab:bbert_parameters}:
The rationale for selecting these models is to provide a diverse assessment across different classification paradigms. SVM, when combined with $n$-grams, is often effective in capturing surface-level patterns whereas BanglaBERT excels in understanding deeper linguistic and contextual nuances, often yielding superior results for minority class predictions, as indicated by previous research \cite{bert_svm_rationale}.
\begin{table}[H]
\centering
    \begin{tabular}{l c}
        \hline
        \textbf{Hyperparameter} & \textbf{Value} \\
        \hline
        Maximum Sequence Length   & 256 \\
        Learning Rate  & 2e-5 \\
        Number of Epochs      & 6 \\
        \hline
    \end{tabular}
    \caption{BanglaBERT's Hyperparameters}
    \label{tab:bbert_parameters}
\end{table}

\section{Results}
Our results indicate that BDA was able to show comparable results in terms of F1 score, reaching close to the performance of 100\% training data while having access to only 50\% of it. It even beats the 100\% data's benchmark in ABSA Cricket and Bangla Sentiment, using BanglaBERT model. Apart from this, in all clipping ranges we can find noticeable improvements, demonstrating BDA's robustness and generalization capabilities.

\begin{table}[H]
\centering
\small
\begin{tabular*}{\textwidth}{@{\extracolsep{\fill}}l c c c c c c c c c}
\toprule
\multirow{2}{*}{\textbf{Model}} & \multicolumn{3}{c}{\textbf{Normal, $T$ \%}} & \multicolumn{3}{c}{\textbf{Augmented, $T'$ \%}} & \multicolumn{3}{c}{\textbf{Diff. $(T'-T)$ \%}} \\\cline{2-10}
 & \textbf{15} & \textbf{50} & \textbf{100} & \textbf{15} & \textbf{50} & \textbf{100} & \textbf{15} & \textbf{50} & \textbf{100} \\ \midrule
BanglaBERT & 64.86 & 66.91 & 70.73 & 64.97 & 68.10 & 72.03 & 0.11 & 1.19 & 1.30\\ 
SVM & 53.69 & 59.54 & 62.42 & 53.83 & 59.76 & 65.79 & 0.14 & 0.22 & \textbf{3.37}\\ 
\textbf{Average} & 59.28 & 63.23 & 66.58 & 59.40 & 63.93 & 68.91 & 0.12 & 0.70 & 2.34 \\ \hline
\end{tabular*}
\caption{SentNoB Dataset F1 scores}
\label{table:SentNoB Dataset comparison}
\end{table}

\begin{table}[H]
\centering
\small
\begin{tabular*}{\textwidth}{@{\extracolsep{\fill}}l c c c c c c c c c}
\toprule
\multirow{2}{*}{\textbf{Model}} & \multicolumn{3}{c}{\textbf{Normal, $T$ \%}} & \multicolumn{3}{c}{\textbf{Augmented, $T'$ \%}} & \multicolumn{3}{c}{\textbf{Diff. $(T'-T)$ \%}} \\\cline{2-10}
 & \textbf{15} & \textbf{50} & \textbf{100} & \textbf{15} & \textbf{50} & \textbf{100} & \textbf{15} & \textbf{50} & \textbf{100} \\ \midrule
BanglaBERT & 59.61 & 68.98 & 70.41 & 61.94 & 69.39 & 70.56 & \textbf{2.33} & 0.41 & 0.14\\ 
SVM & 41.86 & 50.37 & 54.16 & 42.24 & 50.44 & 54.18 & 0.38 & 0.07 & 0.02\\ 
\textbf{Average} & 50.73 & 59.68 & 62.29 & 52.09 & 59.91 & 62.37 & 1.35 & 0.24 & 0.08 \\ \hline
\end{tabular*}
\caption{BemoC Dataset F1 scores}
\label{table:BemoC Dataset comparison}
\end{table}

\begin{table}[H]
\centering
\small
\begin{tabular*}{\textwidth}{@{\extracolsep{\fill}}l c c c c c c c c c}
\toprule
\multirow{2}{*}{\textbf{Model}} & \multicolumn{3}{c}{\textbf{Normal, $T$ \%}} & \multicolumn{3}{c}{\textbf{Augmented, $T'$ \%}} & \multicolumn{3}{c}{\textbf{Diff. $(T'-T)$ \%}} \\\cline{2-10}
 & \textbf{15} & \textbf{50} & \textbf{100} & \textbf{15} & \textbf{50} & \textbf{100} & \textbf{15} & \textbf{50} & \textbf{100} \\ \midrule
BanglaBERT & 45.12 & 48.95 & 49.08 & 45.55 & 49.19 & 49.19 & 0.43 & 0.24 & 0.11\\ 
SVM & 36.38 & 37.90 & 39.51 & 37.01 & 38.70 & 39.54 & 0.63 & \textbf{0.80} & 0.03\\ 
\textbf{Average} & 40.75 & 43.43 & 44.30 & 41.28 & 43.95 & 44.37 & 0.53 & 0.52 & 0.07 \\ \hline
\end{tabular*}
\caption{Bengali Sentiment Dataset F1 scores}
\label{table:Bengali Sentiment Dataset comparison}
\end{table}

%%%%% Bengali Sentiment 15% er bbert ta bhul %%%%%

\begin{table}[H]
\centering
\small
\begin{tabular*}{\textwidth}{@{\extracolsep{\fill}}l c c c c c c c c c}
\toprule
\multirow{2}{*}{\textbf{Model}} & \multicolumn{3}{c}{\textbf{Normal, $T$ \%}} & \multicolumn{3}{c}{\textbf{Augmented, $T'$ \%}} & \multicolumn{3}{c}{\textbf{Diff. $(T'-T)$ \%}} \\\cline{2-10}
 & \textbf{15} & \textbf{50} & \textbf{100} & \textbf{15} & \textbf{50} & \textbf{100} & \textbf{15} & \textbf{50} & \textbf{100} \\ \midrule
BanglaBERT & 38.68 & 49.27 & 54.83 & 46.98 & 56.05 & 57.45 & \textbf{8.30} & 6.77 & 2.62  \\
SVM & 34.80 & 45.71 & 49.63 & 35.81 & 46.59 & 50.42 & 1.01 & 0.88 & 0.79  \\
Average & 36.74 & 47.49 & 52.23 & 41.39 & 51.32 & 53.93 & 4.65 & 3.83 & 1.70\\ \hline
\end{tabular*}
\caption{ABSA Cricket Dataset F1 scores}
\label{table:ABSA Cricket Dataset comparison}
\end{table}

%%%%% ABSA Cricket %%%%%

\begin{table}[H]
\centering
\small
\begin{tabular*}{\textwidth}{@{\extracolsep{\fill}}l c c c c c c c c c}
\toprule
\multirow{2}{*}{\textbf{Model}} & \multicolumn{3}{c}{\textbf{Normal, $T$ \%}} & \multicolumn{3}{c}{\textbf{Augmented, $T'$ \%}} & \multicolumn{3}{c}{\textbf{Diff. $(T'-T)$ \%}} \\\cline{2-10}
 & \textbf{15} & \textbf{50} & \textbf{100} & \textbf{15} & \textbf{50} & \textbf{100} & \textbf{15} & \textbf{50} & \textbf{100} \\ \midrule
BanglaBERT & 23.62 & 38.01 & 45.73 & 26.14 & 45.59 & 55.29 & 2.52 & 7.58 & \textbf{9.56}  \\
SVM & 23.32 & 28.57 & 33.53 & 25.55 & 31.66 & 36.72 & 2.23 & 3.09 & 3.19  \\
Average & 23.47 & 33.29 & 39.63 & 25.84 & 38.62 & 46.00 & 2.37 & 5.33 & 6.37 \\ \hline
\end{tabular*}
\caption{ABSA Restaurant Dataset F1 scores}
\label{table:ABSA Restaurant Dataset comparison}
\end{table}

\subsection{Observations}
\noindent Our test results indicate how augmentation affects datasets and how well models can generalize as follows:
\begin{enumerate}
    \item \textbf{Effect of Augmentation:} Both models generally benefit from data augmentation, especially as the dataset size increases. Augmentation introduces variation in the dataset allowing the model to generalize better. We see across all clipping ranges of the SentNoB and the ABSA Restaurant dataset, the F1 score improved with augmentation, along with other datasets showing improvement in the majority of the cases. Also, the highest amount of improvements can be seen in the 15\% range as well, which is an indicator of BDA's ability to solve data scarcity.
    \item \textbf{Model Comparison:} BanglaBERT generally shows a more robust improvement in F1 scores with augmentation across different datasets and sizes compared to SVM. For instance, in the BemoC dataset, BanglaBERT consistently outperforms SVM variants with or without augmentation. This is due to BanglaBERT's capabilities as a BERT model, which is better at capturing contextual information from augmented text. On the other hand, SVM's performance improves with a larger amount of data but falls short in comparison to BanglaBERT. A detailed study of each model's performance on different clipping ranges is shown in Table \ref{compare_normal_bda_all}.
    \item \textbf{Impact of Dataset Size:} With smaller datasets, augmentation functions similarly to oversampling, producing augmented texts that are only slightly different from the original data. In such cases, memorization can actually enhance performance, improving F1 scores even when using as little as 15\% of the dataset. However, as the dataset size increases, such as when 50\% of the data is used, the model benefits more from the greater diversity and reduced memorization, allowing for better generalization. For example, in the BemoC dataset, 50\% augmented data using BanglaBERT achieved a higher F1 score compared to 100\% of the non-augmented data. Similarly, other datasets like SentNoB, Bengali Sentiment, and ABSA showed comparable trends, highlighting the general benefit of providing more samples for BDA.

    \item \textbf{Lexical Variation in Longer Sentences:} In longer sentences, altering a few words constitutes a smaller fraction of the overall content compared to shorter sentences. This means that longer texts can tolerate more modifications without significantly impacting their original meaning or class label. Consequently, we can introduce more lexical variations by increasing the augmentation level ($n$), the number of altered words, when applying rule-based methods like Synonym Replacement (SR) and Random Swap (RS) to longer sentences. This approach allows us to enrich the dataset with diverse linguistic expressions while maintaining the semantic integrity necessary for accurate classification. 

    \item \textbf{Filtering Low-Quality Augmented Texts:} An additional challenge in Bangla text augmentation is that some sentences offer limited possibilities for rephrasing without altering their class label. This constraint can lead to augmented texts that are nearly identical to the originals, providing minimal variation. Such low-quality outputs need to be filtered out because they do not contribute meaningful diversity to the dataset. To maintain lexical variety while preserving semantic similarity, we set specific thresholds for filtering the generated texts from BDA.
\end{enumerate}

\subsection{Preserving Class Labels in Augmentation}

\begin{figure}[H]
    \centering
    \includegraphics[width=10cm]{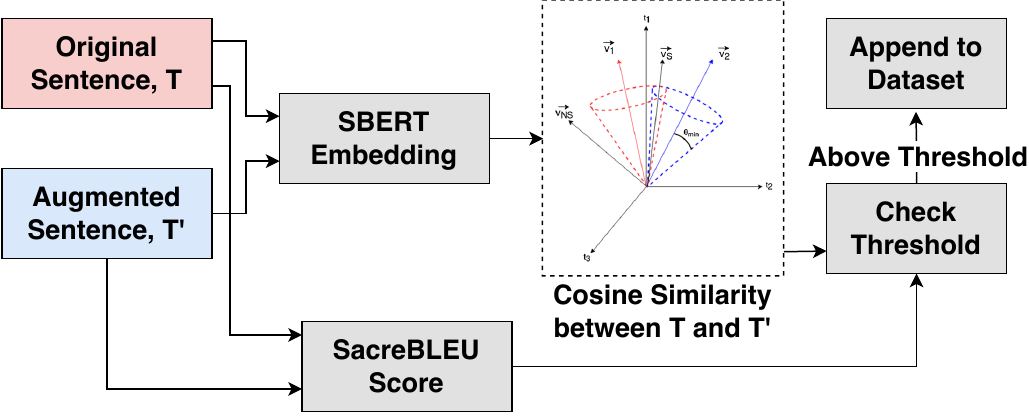}
    \caption{Text filtering process}
    \label{fig:filtering_pipeline}
\end{figure}

A key aspect of high-quality text augmentation is maintaining the integrity of the class or label of the data. The lexical variation in BDA can be controlled through its pipeline parameters, but instead of overly constraining these changes, we employ a filtering process to ensure the generated texts retain their original meaning, as illustrated in Figure~\ref{fig:filtering_pipeline}. This approach prevents adding misleading augmented texts to the dataset. In the filtering process, both the original and augmented sentences are converted into vector representations using a multilingual Sentence-BERT model \cite{sentencebert}. We then compute the cosine similarity to assess how closely the meanings of the two texts align. If the similarity score exceeds a predefined threshold, the augmented text proceeds to the second phase, where we check lexical similarity. For this, we use SacreBLEU to measure $n$-gram overlaps between the sentences. A lower BLEU score indicates fewer shared words between the original and augmented texts. Only augmented texts that meet both the meaning and lexical similarity thresholds are added to the dataset. The thresholds are dataset-specific, balancing lexical diversity with valid augmented samples to ensure meaningful augmentations.

\begin{figure*}
    \centering
    \begin{minipage}[b]{0.18\textwidth}
        \centering
        \includegraphics[width=\textwidth]{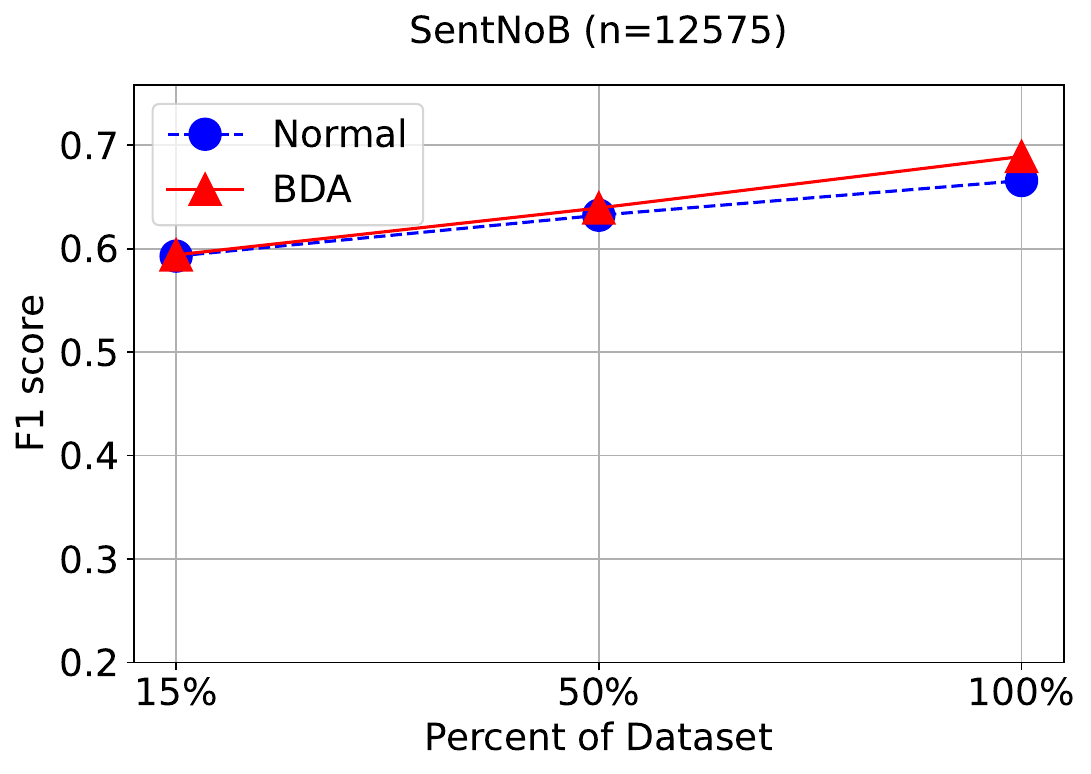}
        % \caption{SentNoB}
        \label{fig:sentnob}
    \end{minipage}
    \hfill
    \begin{minipage}[b]{0.18\textwidth}
        \centering
        \includegraphics[width=\textwidth]{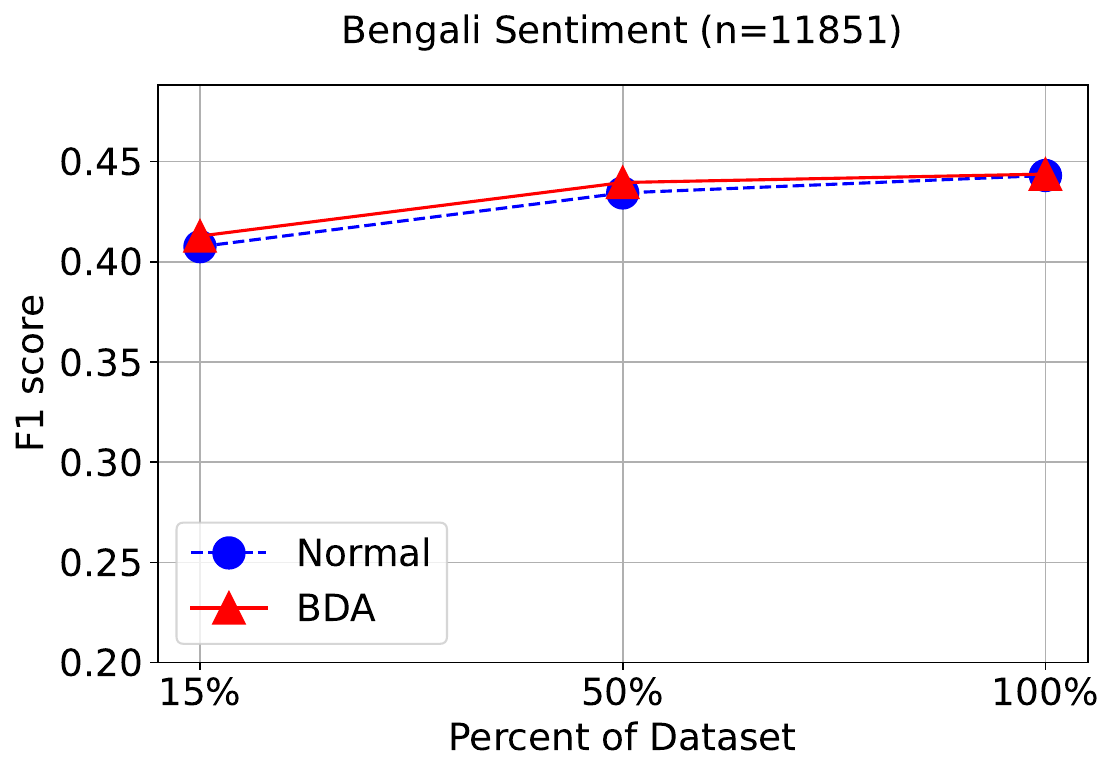}
        % \caption{SentNoB}
        \label{fig:bnsenti}
    \end{minipage}
    \hfill
    \begin{minipage}[b]{0.18\textwidth}
        \centering
        \includegraphics[width=\textwidth]{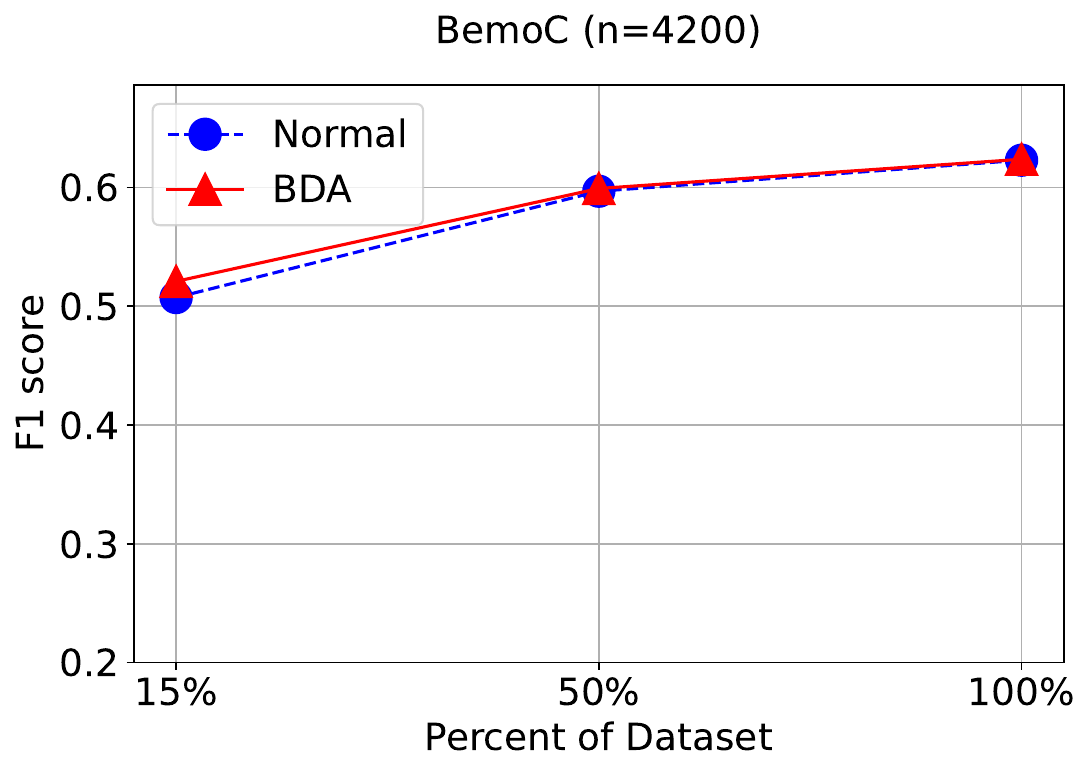}
        % \caption{BEmoC}
        \label{fig:BMOC}
    \end{minipage}
    \hfill
    \begin{minipage}[b]{0.18\textwidth}
        \centering
        \includegraphics[width=\textwidth]{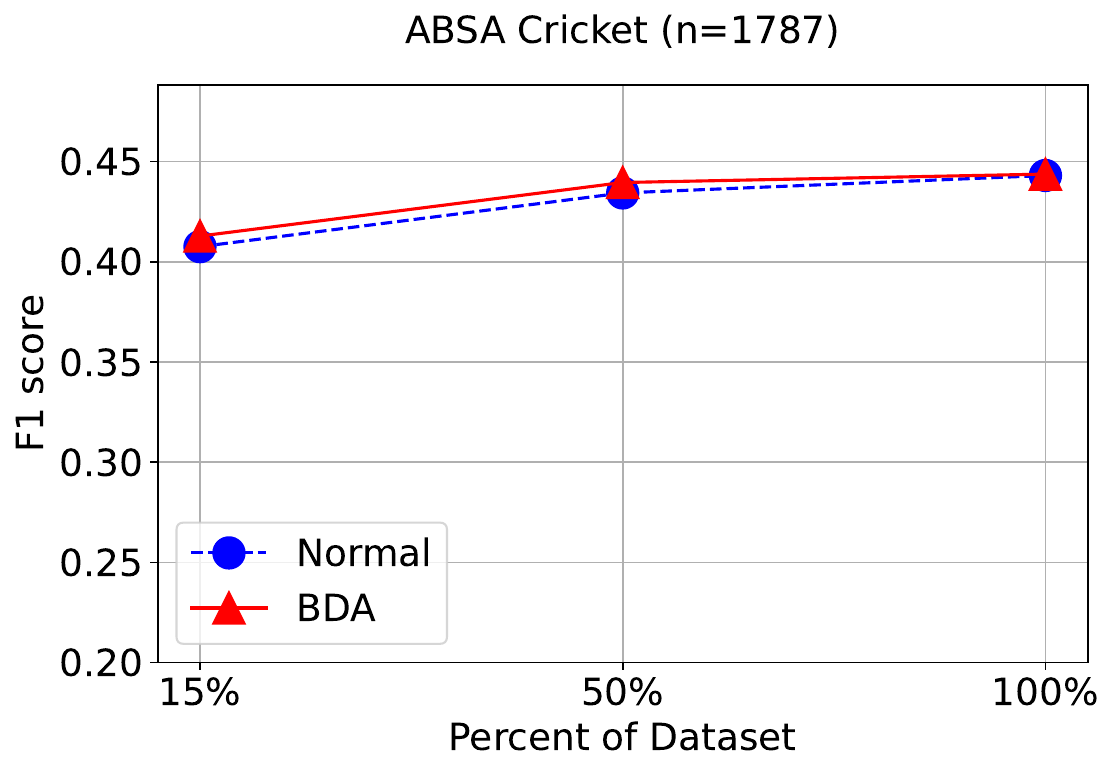}
        % \caption{ABSA C.}
        \label{fig:absa_cricket}
    \end{minipage}
    \hfill
    \begin{minipage}[b]{0.18\textwidth}
        \centering
        \includegraphics[width=\textwidth]{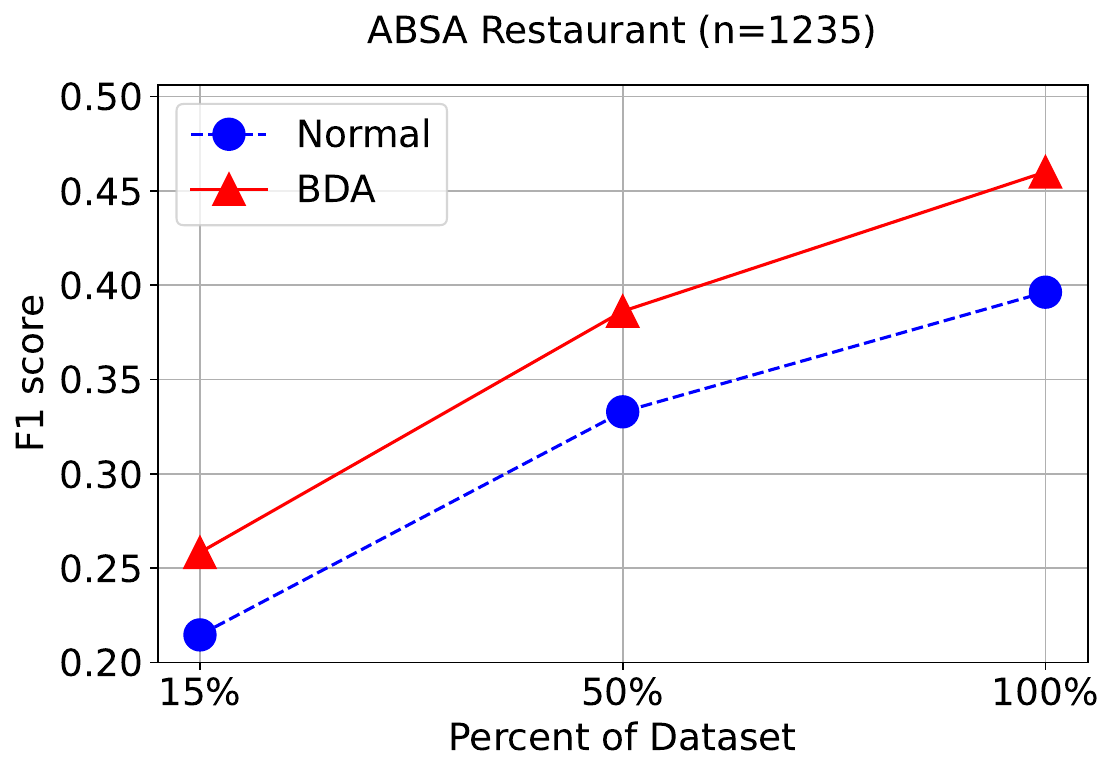}
        % \caption{ABSA R.}
        \label{fig:absa_restaurant}
    \end{minipage}
    \caption{Improvement of F1 scores across all datasets}
    \label{fig:improvement_sizes}
\end{figure*}

\subsection{Ablation Study: Decomposing BDA}

The promising results obtained thus far may lead one to speculate that only a specific part of the BDA pipeline is primarily responsible for the performance gains. To assess the independent contributions of each operation, we isolate and evaluate them individually. We examine all four augmentation methods---Synonym Replacement (SR), Random Swap (RS), Back-Translation (BT), and Paraphrasing (PP)---across all five datasets. The results reveal that each of these operations plays a role in enhancing performance.

\begin{figure}[h]
    \centering
    \includegraphics[width=7cm]{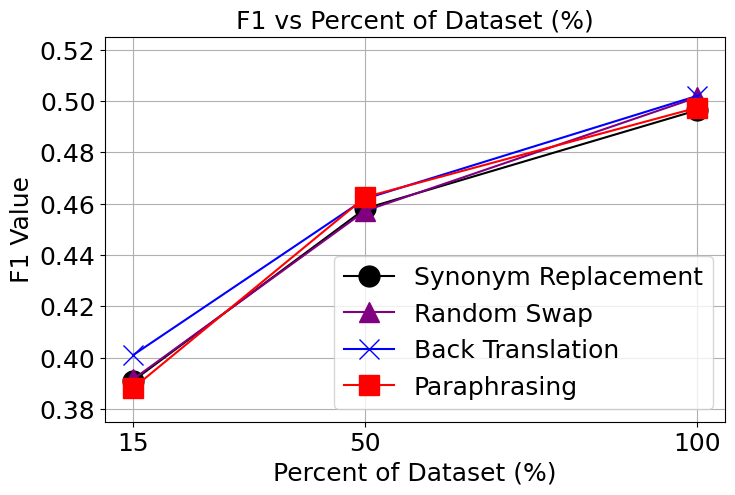}
    \caption{Performance Comparison of Each Augmentation Method}
    \label{fig:all_F1}
\end{figure}

Lexical and semantic analyses, as shown in Table~\ref{table:quality_comparison}, reveal that rule-based approaches—namely Synonym Replacement (SR) and Random Swap (RS) achieve higher SacreBLEU scores compared to Back Translation (BT) or Paraphrasing (PP), while retaining comparable levels of meaning. This occurs because Back-Translation and Paraphrasing alter the complete structure of sentences, whereas Synonym Replacement and Random Swap depend primarily on the number of words selected for alteration based on provided parameters. Among these methods, Random Swap retains the highest semantic similarity. This is likely because, although the Sentence-BERT model \cite{sentencebert} incorporates positional encoding to capture word order, Random Swap only involves swapping existing words without introducing new ones; therefore, the underlying semantics remain largely unchanged.

To study the impact in terms of F1 score, we evaluated the average F1 across all datasets. As illustrated in Figure~\ref{fig:all_F1}, there is no single augmentation method that outperforms others in all scenarios; instead, each method contributes uniquely to generating data variations. The following guidelines outline the most suitable augmentation methods for different types of datasets:

\begin{table}[h]
\centering
\small
\begin{tabular}{l c c}
\hline
\multirow{2}{*}{\textbf{Method}} & \multicolumn{2}{c}{\textbf{Similarity \%}} \\ \cline{2-3}
 & \textbf{Lexical} & \textbf{Semantic} \\ \hline
SR & 59.61 & 83.86 \\ 
RS & 48.57 & 94.06 \\ 
BT & 11.29 & 79.15  \\ 
PP & 15.40 & 86.97 \\ \hline
\end{tabular}
\caption{Quality assessment of Augmented Texts}
\label{table:quality_comparison}
\end{table}

\begin{table}[H]
    \centering
    \begin{tabular}{c c}
    \hline
    \textbf{Method} & \textbf{Ideal Dataset} \\ \hline
    SR & Small to Medium size \\
    RS & Large, Saturated \\ 
    BT & Small and Highly Constrained \\ 
    PP & Versatile Across Dataset Sizes \\ \hline
    
    \end{tabular}
    \caption{Best performing augmentation method according to dataset properties}
    
\end{table}

\section{Discussion and Analysis}

Our results demonstrate that Bangla Data Augmentation (BDA) can improve model performance on specific datasets. However, the extent of improvement depends on the characteristics of the dataset and how effectively the training set is augmented to better represent the test set. The experimental results highlight several key insights into Bangla text augmentation:

\begin{enumerate} 
    \item \textbf{Transformer-based augmentation pipelines generate more natural augmented texts.} This is likely because transformers modify entire sentences rather than individual tokens, leading to greater lexical diversity and richer contextual variations.
    
    \item \textbf{Model-based and rule-based augmentation methods show comparable performance.} F1 scores differ by only 1-2\% across datasets, suggesting both methods can be effective depending on the context. Transformer-based methods offer more diverse and contextually relevant augmentations, while rule-based methods are faster and simpler. In some cases, BanglaBERT outperforms SVM, as seen in Tables \ref{ABSA_Cricket_all} and \ref{ABSA_Restaurant_all}, due to the complexity of the ABSA task, which requires identifying specific aspects or entities in the text.
    
    \item \textbf{No single augmentation method is universally optimal.} The best technique depends on the dataset, task, and model architecture, as shown in Figure \ref{fig:all_F1}. This suggests the need for a hybrid approach that combines different augmentation strategies for optimal results.
\end{enumerate}

\section{Conclusion}
\label{sec:bibtex}
In this work, we addressed the deficiency of effective data augmentation tools and techniques in natural language processing (NLP) for resource-scarce languages like Bangla by developing an augmentation framework called BDA. We experimented with several well-known augmentation techniques previously applied in other languages and incorporated a combination of transformer-based and rule-based approaches to generate a wide array of high-quality synthetic samples with semantically equivalent variations. A novel filtering mechanism was also introduced to enhance the effectiveness of the augmentation.

Our experimental results demonstrated that on smaller datasets, where data diversity is lacking, the BDA framework substantially improved performance by mitigating some of the limitations posed by data scarcity. BDA outperformed the baseline (non-BDA optimized dataset) across five different datasets by increasing the size and diversity of the dataset with high-quality synthetic samples. However, on datasets with the full amount of original data (100\%), performance improvements were very marginal and sometimes even deteriorated. As the scale of data increased, we observed that performance degraded, likely because the augmented samples introduced more noise instead of adding meaningful variety.

\section{Limitations}
One limitation of our framework is its reduced effectiveness in augmenting noisy texts that contain spelling errors or out-of-vocabulary (OOV) words. The Random Swap (RS) method is an exception, as it remains unaffected by such noise due to its meaning-independent nature—it simply swaps word indices without semantic consideration. In contrast, Synonym Replacement (SR), Back Translation (BT), and Paraphrasing (PP) require a certain level of semantic understanding, which hinders their performance on noisy texts. Additionally, informal word structures, such as the growing trend of Banglish (Bangla written in English characters) on social media, pose significant challenges. The Back Translation, Paraphrasing, and Synonym Replacement techniques within BDA are not applicable to such datasets, necessitating the development of specialized methods. Currently, BDA can only augment these texts through Random Swapping.

\section{Future Work}

This study opens several avenues for future research and improvement, with a primary focus on enhancing the sentence similarity model used for filtering augmented data. Although the Sentence-BERT model performs adequately, implementing a more advanced similarity model could yield better results in preventing label flipping of classification datasets in augmented texts. Furthermore, data augmentation is particularly effective in few-shot learning scenarios, where limited training instances are available for each label (e.g., 5-shot or 10-shot settings). Ensuring privacy in sensitive datasets is another critical area. Developing methods to mask sensitive content and selectively augment non-sensitive portions can help protect privacy while still leveraging the benefits of data augmentation.

\appendix
\section{Extended Results}
\label{sec:appendix}
The following tables give an in-depth performance of BDA framework across the discussed 5 dataset, on 3 clipping ranges (15\%, 50\% and 100\%). We used BanglaBERT and variants of SVM as models for evaluation. The F1 scores are reported in order to verify the actual improvement in the robustness of the model, as it only improves if the model is truly getting better at identifying all classes. The tables \ref{SentNoB_all}-\ref{ABSA_Restaurant_all} demonstrate the F1 performance after augmentation using BDA using each individual method: Synonym Replacement, Random Swap, Back-Translation, and Paraphrasing on all the datasets.

\begin{table}
\centering
\resizebox{\textwidth}{!}{%
\begin{tabular}{lcccccccccccc}
\hline
\multirow{2}{*}{\textbf{Model}} & \multicolumn{3}{c}{\textbf{Syn. Replacement}} & \multicolumn{3}{c}{\textbf{Random Swap}} & \multicolumn{3}{c}{\textbf{Back-translation}} & \multicolumn{3}{c}{\textbf{Paraphrasing}} \\ \cline{2-13} 
 & 15\% & 50\% & 100\% & 15\% & 50\% & 100\% & 15\% & 50\% & 100\% & 15\% & 50\% & 100\% \\ \hline
BanglaBERT & 49.11 & 57.13 & 60.39 & 49.36 & 57.76 & 61.06 & 49.45 & 57.83 & 61.33 & 46.22 & 57.93 & 60.75 \\
Unigram (U)& 39.94 & 45.85 & 50.08 & 39.57 & 45.43 & 49.04 & 41.28 & 46.26 & 49.15 & 38.80 & 46.89 & 49.08 \\
Bigram (B)& 33.16 & 39.25 & 44.28 & 33.10 & 38.89 & 43.59 & 33.58 & 40.25 & 44.89 & 32.82 & 39.69 & 44.39 \\
Trigram (T)& 23.46 & 29.68 & 35.18 & 23.60 & 31.27 & 35.28 & 23.66 & 30.22 & 36.13 & 23.66 & 30.05 & 35.73 \\
U+B & 39.68 & 45.44 & 49.25 & 39.86 & 45.30 & 50.61 & 40.75 & 45.81 & 50.02 & 38.55 & 45.37 & 49.08 \\
B+T & 32.34 & 38.48 & 43.56 & 32.67 & 38.47 & 43.04 & 33.36 & 39.22 & 44.40 & 35.77 & 39.60 & 44.26 \\
U+B+T & 40.03 & 45.76 & 50.93 & 40.18 & 45.76 & 51.34 & 41.22 & 47.48 & 51.44 & 39.41 & 45.91 & 50.23 \\
Char 2-gram (C2)& 39.63 & 46.01 & 47.54 & 39.20 & 45.90 & 48.72 & 40.59 & 45.47 & 48.05 & 38.74 & 45.66 & 48.04 \\
Char 3-gram (C3)& 40.90 & 46.92 & 50.13 & 40.34 & 46.89 & 50.95 & 41.70 & 47.79 & 50.41 & 40.39 & 47.06 & 50.50 \\
Char 4-gram (C4)& 39.43 & 46.53 & 50.57 & 40.13 & 46.21 & 51.50 & 40.23 & 46.79 & 50.95 & 39.01 & 46.72 & 50.25 \\
Char 5-gram (C5)& 38.47 & 45.32 & 49.41 & 38.62 & 45.23 & 50.37 & 39.39 & 46.14 & 50.91 & 37.87 & 45.93 & 49.48 \\
C2+C3 & 40.76 & 47.68 & 50.09 & 39.96 & 47.19 & 50.91 & 41.77 & 47.52 & 50.79 & 39.79 & 47.87 & 51.03 \\
C3+C4 & 39.94 & 47.57 & 51.11 & 40.25 & 47.48 & 51.71 & 40.92 & 47.58 & 51.69 & 39.49 & 47.89 & 50.82 \\
C4+C5 & 39.27 & 46.44 & 50.42 & 39.46 & 45.96 & 51.19 & 40.13 & 46.74 & 51.34 & 38.82 & 46.99 & 49.81 \\
C2+C3+C4 & 40.20 & 47.62 & 50.97 & 39.94 & 47.50 & 51.04 & 41.18 & 47.80 & 51.99 & 39.49 & 47.72 & 51.31 \\
C3+C4+C5 & 39.26 & 47.33 & 51.41 & 39.67 & 46.95 & 51.56 & 40.36 & 47.66 & 51.62 & 39.12 & 47.95 & 51.19 \\
C2+C3+C4+C5 & 39.60 & 47.21 & 51.23 & 39.88 & 47.06 & 51.42 & 40.87 & 47.86 & 51.63 & 39.27 & 48.30 & 51.38 \\
U+B+C3+C4+C5 & 39.84 & 46.75 & 51.20 & 39.97 & 46.30 & 51.92 & 40.85 & 47.60 & 51.96 & 40.01 & 47.12 & 51.14 \\
U+B+C2+C3+C4+C5 & 40.01 & 46.98 & 51.14 & 40.26 & 46.48 & 51.84 & 41.48 & 47.78 & 51.91 & 40.02 & 47.86 & 51.41 \\
U+B+T+C2+C3+C4+C5 & 39.66 & 47.02 & 51.64 & 40.10 & 46.66 & 52.30 & 40.94 & 47.32 & 51.65 & 39.83 & 47.59 & 52.06 \\
Embeddings (E)& 42.39 & 45.37 & 45.75 & 42.59 & 45.22 & 47.42 & 43.42 & 44.43 & 46.37 & 42.61 & 45.48 & 46.03 \\
U+B+C2+C3+C4+C5+E & 41.19 & 48.53 & 52.52 & 41.14 & 48.64 & 53.08 & 42.80 & 48.43 & 52.79 & 41.48 & 49.17 & 52.75 \\
U+B+T+C2+C3+C4+C5+E & 41.03 & 48.67 & 52.81 & 40.88 & 49.00 & 53.22 & 42.43 & 48.42 & 53.05 & 41.60 & 49.08 & 53.15 \\
 \hline
\end{tabular}%
}
\caption{Average F1 scores of all datasets for Each method }
\label{avg_method}
\end{table}

\begin{table}
\resizebox{\textwidth}{!}{%
\begin{tabular}{lccccccccccccccc}
\hline
\multirow{2}{*}{\textbf{Model}} & \multicolumn{3}{c}{\textbf{SentNoB}} & \multicolumn{3}{c}{\textbf{BemoC}} & \multicolumn{3}{c}{\textbf{Bengali Sentiment}} & 
\multicolumn{3}{c}{\textbf{ABSA Cricket}} & 
\multicolumn{3}{c}{\textbf{ABSA Restaurant}} \\ \cline{2-16} 
 & 15\% & 50\% & 100\% & 15\% & 50\% & 100\% & 15\% & 50\% & 100\% & 15\% & 50\% & 100\% & 15\% & 50\% & 100\% \\ \hline
BanglaBERT & 64.97 & 68.10 & 72.04 & 61.94 & 69.39 & 70.55 & 47.74 & 49.19 & 49.08 & 41.90 & 56.05 & 57.45 & 

27.98& 45.58 & 55.28 \\
Unigram (U) & 54.49 & 60.93 & 66.61 & 42.48 & 50.26 & 54.66 & 37.43 & 38.96 & 40.04 & 37.71 & 46.92 & 50.77 & 27.38 & 33.49 & 34.59 \\
Bigram (B) & 46.79 & 54.19 & 61.65 & 29.81 & 38.00 & 42.74 & 34.41 & 36.78 & 37.99 & 31.75 & 43.12 & 47.21 & 23.06 & 25.50 & 31.86 \\
Trigram (T) & 31.66 & 43.70 & 54.06 & 11.54 & 21.69 & 25.49 & 25.79 & 28.44 & 30.42 & 30.37 & 37.77 & 42.40 & 18.61 & 19.93 & 25.53 \\
U+B & 55.93 & 60.90 & 67.71 & 43.21 & 51.27 & 55.82 & 37.05 & 37.87 & 39.40 & 35.81 & 44.96 & 50.50 & 26.55 & 32.40 & 35.29 \\
B+T & 46.86 & 54.24 & 62.03 & 29.85 & 37.72 & 43.22 & 33.42 & 35.70 & 36.53 & 34.75 & 42.48 & 46.28 & 22.82 & 24.57 & 31.03 \\
U+B+T & 56.12 & 60.77 & 67.19 & 43.25 & 51.08 & 55.94 & 37.08 & 38.33 & 39.97 & 37.19 & 46.80 & 52.29 & 27.42 & 34.18 & 39.55 \\
Char 2-gram (C2) & 52.75 & 57.99 & 61.46 & 39.78 & 47.96 & 50.90 & 37.78 & 38.19 & 39.03 & 38.11 & 49.03 & 51.63 & 29.28 & 35.63 & 37.42 \\
Char 3-gram (C3) & 56.83 & 60.77 & 66.44 & 45.10 & 52.69 & 55.73 & 39.08 & 39.77 & 40.48 & 37.35 & 48.60 & 52.08 & 25.80 & 34.00 & 37.74 \\
Char 4-gram (C4) & 56.14 & 61.35 & 67.63 & 44.98 & 53.90 & 57.61 & 38.42 & 40.08 & 40.99 & 34.92 & 46.93 & 51.19 & 24.03 & 30.55 & 36.65 \\
Char 5-gram (C5) & 55.10 & 61.94 & 67.70 & 42.77 & 53.12 & 57.52 & 36.51 & 39.70 & 41.04 & 35.86 & 46.37 & 50.12 & 22.70 & 27.13 & 33.84 \\
C2+C3 & 55.75 & 61.67 & 67.25 & 44.78 & 52.70 & 56.02 & 38.65 & 39.68 & 40.36 & 37.81 & 48.98 & 52.47 & 25.84 & 34.80 & 37.43 \\
C3+C4 & 56.94 & 61.69 & 67.94 & 45.26 & 54.26 & 57.82 & 38.80 & 40.30 & 40.91 & 35.29 & 48.01 & 51.86 & 24.46 & 33.88 & 38.12 \\
C4+C5 & 56.36 & 62.48 & 68.25 & 44.82 & 54.22 & 58.42 & 37.36 & 40.02 & 41.56 & 35.35 & 47.02 & 49.58 & 23.22 & 28.93 & 35.62 \\
C2+C3+C4 & 56.24 & 62.01 & 68.47 & 45.45 & 54.26 & 57.78 & 38.52 & 40.09 & 40.97 & 35.75 & 48.17 & 52.03 & 25.05 & 33.79 & 37.36 \\
C3+C4+C5 & 56.32 & 62.37 & 68.91 & 45.26 & 54.60 & 59.15 & 37.81 & 39.92 & 41.42 & 35.06 & 47.93 & 50.79 & 23.57 & 32.54 & 36.94 \\
C2+C3+C4+C5 & 56.32 & 62.29 & 68.94 & 45.71 & 54.84 & 59.06 & 37.76 & 39.87 & 41.36 & 35.18 & 48.40 & 50.79 & 24.53 & 32.62 & 36.92 \\
U+B+C3+C4+C5 & 54.42 & 61.92 & 67.99 & 45.57 & 53.83 & 56.71 & 37.31 & 39.52 & 40.98 & 36.08 & 47.00 & 52.75 & 27.47 & 32.46 & 39.34 \\
U+B+C2+C3+C4+C5 & 54.87 & 62.11 & 68.20 & 45.78 & 53.62 & 56.62 & 37.42 & 39.64 & 40.82 & 36.22 & 47.60 & 52.55 & 27.92 & 33.42 & 39.68 \\
U+B+T+C2+C3+C4+C5 & 55.01 & 62.10 & 68.50 & 45.89 & 54.06 & 57.24 & 37.40 & 39.25 & 41.20 & 35.61 & 47.92 & 52.48 & 26.75 & 32.42 & 40.14 \\
Embeddings (E) & 55.54 & 54.40 & 53.27 & 50.44 & 53.90 & 56.62 & 38.86 & 38.72 & 38.78 & 38.03 & 45.58 & 45.54 & 30.88 & 33.02 & 37.75 \\
U+B+C2+C3+C4+C5+E & 56.44 & 62.33 & 68.54 & 48.86 & 55.58 & 58.23 & 38.76 & 40.32 & 41.66 & 36.58 & 49.17 & 53.34 & 27.61 & 36.07 & 42.15 \\
U+B+T+C2+C3+C4+C5+E & 57.36 & 62.50 & 68.66 & 48.66 & 56.05 & 58.71 & 38.62 & 40.30 & 41.87 & 35.76 & 49.60 & 53.18 & 27.04 & 35.52 & 42.86 \\ \hline
\end{tabular}%
}
\caption{Table of average F1 scores for each dataset after using BDA}
\label{all_scores_all_dataset_summarized}
\end{table}

\begin{table}
\resizebox{\textwidth}{!}{%
\begin{tabular}{lcccccccccccc}
\hline
\multirow{2}{*}{\textbf{Model}} & \multicolumn{3}{c}{\textbf{Syn. Replacement}} & \multicolumn{3}{c}{\textbf{Random Swap}} & \multicolumn{3}{c}{\textbf{Back-translation}} & \multicolumn{3}{c}{\textbf{Paraphrasing}} \\ \cline{2-13} 
 & 15\% & 50\% & 100\% & 15\% & 50\% & 100\% & 15\% & 50\% & 100\% & 15\% & 50\% & 100\% \\ \hline
BanglaBERT & 63.86& 67.31& 70.45& 65.40& 68.49& 71.31& 65.40& 68.39 & 72.04 & 65.21& 68.22&              74.33\\
Unigram (U) & 55.24& 61.93& 67.30& 53.62& 60.11& 66.81& 55.22& 60.61 & 66.34 & 53.87& 61.05& 65.99\\
Bigram (B) & 47.21& 54.16& 62.01& 46.39& 53.95& 61.89& 47.19& 54.76 & 61.70 & 46.35& 53.90& 61.01\\
Trigram (T) & 31.41& 43.58& 53.58& 32.30& 43.67& 54.44& 31.40& 43.80 & 54.20 & 31.53& 43.74& 54.01\\
U+B & 56.21& 61.04& 67.10& 55.59& 60.43& 68.82& 56.23& 60.84 & 66.93 & 55.72& 61.29& 67.99\\
B+T & 46.80& 53.46& 62.41& 47.20& 54.29& 62.35& 46.79& 54.64 & 61.68 & 46.62& 54.55& 61.68\\
U+B+T & 56.35& 60.83& 66.94& 56.10& 60.77& 68.06& 56.34& 60.63 & 66.77 & 55.68& 60.84& 67.01\\
Char 2-gram (C2) & 53.51& 57.33& 61.26& 51.85& 57.55& 62.51& 53.4& 59.76 & 61.06 & 52.14& 57.33& 61.00\\
Char 3-gram (C3) & 57.42& 60.68& 65.91& 55.93& 60.72& 66.98& 57.44& 61.39 & 65.89 & 56.54& 60.29& 66.98\\
Char 4-gram (C4) & 55.66& 61.08& 67.84& 57.11& 60.46& 67.87& 55.67& 61.89 & 66.82 & 56.15& 61.96& 67.99\\
Char 5-gram (C5) & 55.19& 61.60& 67.85& 55.06& 61.25& 68.19& 55.17& 62.39 & 67.78 & 54.96& 62.53& 67.01\\
C2+C3 & 56.40& 61.61& 66.87& 54.38& 61.62& 67.66& 56.41& 61.83 & 66.48 & 55.81& 61.61& 67.99\\
C3+C4 & 56.86& 61.43& 68.03& 57.22& 61.57& 68.18& 56.66& 61.80 & 67.57 & 56.81& 61.96& 68.01\\
C4+C5 & 56.55& 62.68& 68.61& 56.45& 62.26& 68.39& 56.65& 62.41 & 68.00 & 55.87& 62.55& 67.99\\
C2+C3+C4 & 56.02& 61.81& 68.02& 56.47& 61.09& 68.28& 56.01& 62.73 & 68.58 & 56.44& 62.40& 68.99\\
C3+C4+C5 & 56.01& 62.63& 69.39& 57.03& 62.16& 69.24& 56.01& 62.16 & 68.03 & 56.21& 62.54& 68.98\\
C2+C3+C4+C5 & 56.52& 62.41& 69.19& 56.12& 61.42& 69.14& 56.54& 62.10 & 68.45 & 56.13& 63.23& 69.02\\
U+B+C3+C4+C5 & 53.96& 61.21& 67.85& 55.73& 61.62& 68.70& 53.99& 62.37 & 67.40 & 54.04& 62.48& 67.99\\
U+B+C2+C3+C4+C5 & 54.87& 61.46& 68.23& 55.35& 61.82& 68.74& 54.85& 62.51 & 67.85 & 54.38& 62.64& 67.99\\
U+B+T+C2+C3+C4+C5 & 54.91& 61.39& 68.67& 55.17& 61.86& 68.84& 54.92& 62.36 & 67.48 & 55.05& 62.77& 69.01\\
Embeddings (E) & 55.67& 54.86& 55.18& 56.14& 54.94& 55.86& 53.88& 53.65 & 54.05 & 56.49& 54.17& 48.01\\
U+B+C2+C3+C4+C5+E & 56.06& 61.96& 68.40& 56.30& 61.99& 69.30& 56.61& 62.50 & 67.45 & 56.79& 62.87& 68.99\\
U+B+T+C2+C3+C4+C5+E & 56.77& 62.02& 69.08& 57.35& 62.14& 68.71& 57.65& 62.80 & 67.85 & 57.65& 63.02& 68.99\\ \hline
\end{tabular}%
}
\caption{F1 scores of SentNoB dataset for Each method}
\label{SentNoB_all}
\end{table}

\begin{table}
\resizebox{\textwidth}{!}{%
\begin{tabular}{lcccccccccccc}
\hline
\multirow{2}{*}{\textbf{Model}} & \multicolumn{3}{c}{\textbf{Syn. Replacement}} & \multicolumn{3}{c}{\textbf{Random Swap}} & \multicolumn{3}{c}{\textbf{Back-translation}} & \multicolumn{3}{c}{\textbf{Paraphrasing}} \\ \cline{2-13} 
 & 15\% & 50\% & 100\% & 15\% & 50\% & 100\% & 15\% & 50\% & 100\% & 15\% & 50\% & 100\% \\ \hline
BanglaBERT & 60.99& 69.01& 70.47 & 64.11& 69.61& 70.61 & 60.87& 69.63& 70.62& 61.78& 69.31& 70.52 \\
Unigram (U) & 42.16& 49.70& 53.87 & 41.92& 49.75& 54.20 & 45.02& 50.51& 55.14& 40.82& 51.08& 55.44 \\
Bigram (B) & 29.25& 36.24& 41.50 & 30.04& 37.64& 42.41 & 30.02& 38.76& 41.90& 29.94& 39.38& 45.15 \\
Trigram (T) & 11.23& 18.67& 24.07 & 10.85& 27.07& 24.44 & 12.04& 20.26& 26.37& 12.03& 20.77& 27.09 \\
U+B & 42.66& 50.70& 55.32 & 43.06& 51.43& 56.44 & 44.68& 51.91& 55.67& 42.45& 51.04& 55.83 \\
B+T & 29.49& 36.40& 41.97 & 29.25& 36.75& 42.55 & 30.34& 38.09& 43.02& 30.33& 39.64& 45.36 \\
U+B+T & 42.42& 50.13& 55.47 & 42.74& 51.01& 55.81 & 45.17& 52.03& 56.15& 42.67& 51.13& 56.31 \\
Char 2-gram (C2) & 38.43& 48.45& 50.97 & 39.06& 47.90& 49.86 & 41.57& 47.79& 51.49& 40.05& 47.70& 51.29 \\
Char 3-gram (C3) & 44.37& 52.79& 55.24 & 45.23& 51.71& 55.09 & 45.69& 53.58& 56.40& 45.11& 52.68& 56.18 \\
Char 4-gram (C4) & 45.00& 54.07& 58.18 & 44.94& 54.63& 58.09 & 45.64& 53.81& 56.84& 44.34& 53.07& 57.34 \\
Char 5-gram (C5) & 42.38& 52.71& 57.08 & 43.64& 53.48& 57.12 & 42.33& 53.38& 57.62& 42.74& 52.93& 58.24 \\
C2+C3 & 43.80& 53.55& 55.59 & 44.39& 52.00& 55.53 & 46.06& 52.93& 56.84& 44.89& 52.34& 56.11 \\
C3+C4 & 44.88& 54.17& 57.27 & 45.16& 54.69& 58.18 & 45.94& 54.37& 58.57& 45.06& 53.80& 57.27 \\
C4+C5 & 44.18& 54.07& 58.60 & 44.80& 54.59& 58.35 & 45.74& 54.24& 58.18& 44.57& 53.98& 58.57 \\
C2+C3+C4 & 45.20& 54.40& 57.33 & 45.06& 54.75& 57.26 & 46.38& 54.34& 58.96& 45.17& 53.55& 57.59 \\
C3+C4+C5 & 45.03& 54.22& 59.06 & 45.00& 54.35& 58.78 & 45.25& 55.06& 59.31& 45.75& 54.77& 59.46 \\
C2+C3+C4+C5 & 44.77& 54.62& 58.61 & 46.21& 55.10& 59.07 & 46.07& 54.91& 59.33& 45.80& 54.73& 59.25 \\
U+B+C3+C4+C5 & 45.04& 53.76& 56.28 & 45.45& 52.77& 56.78 & 45.94& 54.77& 56.85& 45.84& 54.02& 56.92 \\
U+B+C2+C3+C4+C5 & 45.07& 53.23& 56.00 & 46.11& 52.28& 56.92 & 45.80& 55.00& 56.79& 46.14& 53.97& 56.78 \\
U+B+T+C2+C3+C4+C5 & 44.91& 54.06& 56.99 & 46.09& 52.91& 57.33 & 46.05& 54.86& 56.67& 46.52& 54.43& 57.98 \\
Embeddings (E) & 49.40& 54.42& 56.15 & 50.23& 53.43& 55.97 & 51.64& 53.04& 56.66& 50.49& 54.72& 57.70 \\
U+B+C2+C3+C4+C5+E & 48.44& 55.57& 57.73 & 47.81& 55.44& 58.53 & 50.18& 56.00& 58.39& 49.03& 55.29& 58.26 \\
U+B+T+C2+C3+C4+C5+E & 47.93& 55.73& 58.10 & 47.61& 56.36& 59.04 & 49.77& 56.27& 58.87& 49.31& 55.83& 58.84 \\ \hline
\end{tabular}%
}
\caption{F1 scores of BemoC dataset for Each method}
\label{BemoC_all}
\end{table}

\begin{table}
\resizebox{\textwidth}{!}{%
\begin{tabular}{lcccccccccccc}
\hline
\multirow{2}{*}{\textbf{Model}} & \multicolumn{3}{c}{\textbf{Syn. Replacement}} & \multicolumn{3}{c}{\textbf{Random Swap}} & \multicolumn{3}{c}{\textbf{Back-translation}} & \multicolumn{3}{c}{\textbf{Paraphrasing}} \\ \cline{2-13} 
 & 15\% & 50\% & 100\% & 15\% & 50\% & 100\% & 15\% & 50\% & 100\% & 15\% & 50\% & 100\% \\ \hline
BanglaBERT & 48.58 & 49.20 & 49.59 & 47.93 & 50.28 & 49.32& 37.68& 48.51 & 48.70& 48.00 & 48.76& 48.79\\
Unigram (U) & 37.73 & 38.14 & 39.38& 37.72 & 38.62 & 39.92& 34.52& 39.87 & 39.87& 36.57 & 39.22& 39.61\\
Bigram (B) & 34.20 & 36.38 & 36.97& 34.49 & 36.66 & 37.73& 25.90& 37.34 & 37.34& 34.43 & 36.74& 37.46\\
Trigram (T) & 25.66 & 28.21 & 30.32& 25.86 & 28.08 & 30.32& 37.11& 29.78 & 29.78& 25.75 & 27.68& 29.89\\
U+B & 37.12 & 37.69 & 38.91& 37.33 & 37.25 & 38.69& 33.50& 39.29 & 39.29& 36.63 & 37.26& 38.98\\
B+T & 32.93 & 35.74 & 35.52& 33.44 & 35.60 & 36.49& 37.28& 35.78 & 35.78& 33.80 & 35.68& 35.54\\
U+B+T & 36.74 & 37.78 & 39.32& 37.20 & 38.54 & 39.37& 36.87& 39.39 & 39.39& 37.08 & 37.60& 39.65\\
Char 2-gram (C2) & 38.63 & 37.99 & 38.95& 38.01 & 37.71 & 38.14& 39.03& 38.38 & 38.38& 37.60 & 38.68& 38.39\\
Char 3-gram (C3) & 39.31 & 39.20 & 40.34& 39.47 & 39.65 & 40.05& 38.61& 40.49 & 40.49& 38.50 & 39.73& 40.12\\
Char 4-gram (C4) & 38.12 & 39.55 & 41.02& 38.98 & 39.59 & 41.40& 36.97& 40.94 & 40.94& 37.98 & 40.26& 40.59\\
Char 5-gram (C5) & 36.54 & 38.56 & 40.35& 36.53 & 39.75 & 40.39& 39.11& 40.91 & 40.91& 36.01 & 39.58& 40.68\\
C2+C3 & 39.20 & 39.46 & 40.59& 38.89 & 39.10 & 40.07& 38.96& 40.40 & 40.40& 37.41 & 39.74& 39.98\\
C3+C4 & 38.64 & 40.00 & 40.24& 39.41 & 39.73 & 40.91& 37.72& 40.50 & 40.50& 38.20 & 40.99& 40.53\\
C4+C5 & 37.33 & 39.41 & 41.06& 37.22 & 39.85 & 40.43& 39.00& 40.78 & 40.78& 37.16 & 40.05& 41.10\\
C2+C3+C4 & 38.53 & 39.54 & 39.98& 38.87 & 39.82 & 40.81& 38.27& 40.67 & 40.67& 37.66 & 40.31& 40.36\\
C3+C4+C5 & 37.80 & 39.31 & 40.98& 38.18 & 39.31 & 41.13& 38.21& 40.86 & 40.86& 37.00 & 40.18& 40.97\\
C2+C3+C4+C5 & 37.85 & 38.98 & 40.83& 38.04 & 39.28 & 40.79& 36.92& 40.92 & 40.92& 36.95 & 40.31& 40.96\\
U+B+C3+C4+C5 & 37.52 & 38.70 & 40.61& 37.17 & 38.72 & 40.74& 37.36& 41.22 & 41.22& 37.63 & 39.43& 40.71\\
U+B+C2+C3+C4+C5 & 37.47 & 39.41 & 40.35& 37.12 & 38.83 & 40.54& 37.32& 41.29 & 41.29& 37.75 & 39.03& 40.51\\
U+B+T+C2+C3+C4+C5 & 37.20 & 38.91 & 40.67& 37.47 & 38.69 & 40.84& 38.90& 40.16 & 41.64& 37.62 & 39.24& 40.98\\
Embeddings (E) & 38.57 & 39.14 & 38.87& 39.13 & 38.98 & 38.18& 39.11& 38.19 & 37.13& 38.85 & 38.59& 37.85\\
U+B+C2+C3+C4+C5+E & 38.48 & 40.16 & 41.15& 38.49 & 40.01 & 40.55& 38.77& 40.96 & 41.39& 38.98 & 40.14& 41.26\\
U+B+T+C2+C3+C4+C5+E & 38.37 & 40.19 & 40.90& 38.40 & 40.12 & 41.00& 36.90& 41.13 & 41.71& 38.94 & 39.77& 41.51\\ \hline
\end{tabular}%
}
\caption{F1 scores of Bengali Sentiment for Each method}
\label{Bn_senti_all}
\end{table}

\begin{table}
\resizebox{\textwidth}{!}{%
\begin{tabular}{lcccccccccccc}
\hline
\multirow{2}{*}{\textbf{Model}} & \multicolumn{3}{c}{\textbf{Syn. Replacement}} & \multicolumn{3}{c}{\textbf{Random Swap}} & \multicolumn{3}{c}{\textbf{Back-translation}} & \multicolumn{3}{c}{\textbf{Paraphrasing}} \\ \cline{2-13} 
 & 15\% & 50\% & 100\% & 15\% & 50\% & 100\% & 15\% & 50\% & 100\% & 15\% & 50\% & 100\% \\ \hline
BanglaBERT & 45.72& 55.85& 57.28& 45.57 & 56.01 & 59.78 & 48.32 & 56.72 & 57.90 & 48.32& 55.60& 54.83\\
Unigram (U) & 37.42& 47.27& 51.69& 37.42 & 46.37 & 49.95 & 39.53 & 47.05 & 52.10 & 39.53& 46.97& 49.35\\
Bigram (B) & 32.05& 43.24& 48.14& 32.05 & 44.41 & 47.26 & 31.60 & 41.82 & 46.55 & 31.60& 43.01& 46.87\\
Trigram (T) & 30.37& 38.04& 42.54& 30.37 & 37.62 & 42.36 & 30.37 & 37.29 & 42.36 & 30.37 & 38.13& 42.36\\
U+B & 36.15& 44.96& 49.53& 36.15 & 45.11 & 52.15 & 36.92 & 45.74 & 51.68 & 34.01 & 44.02& 48.62\\
B+T & 30.37& 43.50& 46.75& 30.37 & 43.16 & 46.17 & 31.64 & 40.63 & 46.31 & 46.62 & 42.64& 45.87\\
U+B+T & 37.32& 45.18& 52.61& 37.32 & 46.64 & 53.26 & 38.82 & 49.46 & 52.06 & 35.29 & 45.90& 51.24\\
Char 2-gram (C2) & 38.47& 50.03& 50.75& 38.47 & 50.58 & 53.32 & 39.28 & 46.80 & 50.35 & 36.23 & 48.70& 52.09\\
Char 3-gram (C3) & 37.09& 47.72& 52.19& 37.09 & 47.60 & 53.10 & 38.15 & 48.40 & 52.12 & 37.07 & 50.70& 50.93\\
Char 4-gram (C4) & 34.41& 47.11& 48.77& 34.41 & 45.59 & 53.22 & 35.97 & 46.90 & 53.82 & 34.91 & 48.12& 48.96\\
Char 5-gram (C5) & 35.91& 46.42& 49.07& 35.91 & 44.66 & 50.53 & 38.02 & 47.13 & 52.12 & 33.60 & 47.28& 48.76\\
C2+C3 & 38.04& 48.51& 51.86& 38.04 & 48.83 & 52.94 & 37.97 & 47.77 & 53.02 & 37.21 & 50.81& 52.06\\
C3+C4 & 35.01& 46.52& 51.42& 35.01 & 47.12 & 52.52 & 36.70 & 48.27 & 53.57 & 34.45 & 50.14& 49.93\\
C4+C5 & 35.04& 46.78& 47.84& 35.04 & 44.75 & 50.53 & 36.71 & 47.62 & 52.15 & 34.63 & 48.92& 47.81\\
C2+C3+C4 & 35.38& 47.71& 51.59& 35.38 & 47.38 & 52.36 & 37.43 & 48.19 & 52.84 & 34.83 & 49.40& 51.32\\
C3+C4+C5 & 34.35& 47.04& 49.39& 34.35 & 46.92 & 50.72 & 37.09 & 47.77 & 53.21 & 34.45 & 50.00& 49.84\\
C2+C3+C4+C5 & 34.77& 47.37& 49.39& 34.77 & 46.83 & 51.00 & 36.74 & 49.34 & 52.18 & 35.26 & 50.07& 50.58\\
U+B+C3+C4+C5 & 35.36& 47.57& 52.96& 35.36 & 45.38 & 53.16 & 38.33 & 47.60 & 53.26 & 34.96 & 47.45& 51.62\\
U+B+C2+C3+C4+C5 & 35.36& 47.92& 52.88& 35.45 & 45.54 & 52.22 & 39.20 & 48.05 & 52.97 & 34.16 & 48.87& 52.12\\
U+B+T+C2+C3+C4+C5 & 35.45& 48.07& 52.76& 36.94 & 46.09 & 53.24 & 37.39 & 48.73 & 52.27 & 37.63 & 48.77& 51.65\\
Embeddings (E) & 36.94& 45.96& 43.84& 36.07 & 44.81 & 47.81 & 40.63 & 45.81 & 43.36 & 35.84 & 45.75& 47.13\\
U+B+C2+C3+C4+C5+E & 36.07& 49.19& 53.63& 35.11 & 50.42 & 52.90 & 38.35 & 48.56 & 53.09 & 35.91 & 48.52& 53.75\\
U+B+T+C2+C3+C4+C5+E & 35.11& 49.23& 53.02& 35.11 & 50.60 & 53.25 & 36.90 & 48.54 & 53.07 & 35.91 & 50.04& 53.39\\ \hline
\end{tabular}%
}
\caption{F1 scores of ABSA Cricket for Each method }
\label{ABSA_Cricket_all}
\end{table}

\begin{table}
\resizebox{\textwidth}{!}{%
\begin{tabular}{lcccccccccccc}
\hline
\multirow{2}{*}{\textbf{Model}} & \multicolumn{3}{c}{\textbf{Syn. Replacement}} & \multicolumn{3}{c}{\textbf{Random Swap}} & \multicolumn{3}{c}{\textbf{Back-translation}} & \multicolumn{3}{c}{\textbf{Paraphrasing}} \\ \cline{2-13} 
 & 15\% & 50\% & 100\% & 15\% & 50\% & 100\% & 15\% & 50\% & 100\% & 15\% & 50\% & 100\% \\ \hline
BanglaBERT & 26.42& 44.30 & 54.14 & 23.78 & 44.39 & 54.35 & 26.22 & 45.91 & 57.37 & 28.15& 47.74& 55.28\\
Unigram (U) & 27.16& 32.23 & 37.92 & 27.17 & 32.32 & 34.61 & 28.94 & 33.27 & 30.82 & 

26.24& 36.15& 35.00\\
Bigram (B) & 23.08& 26.22 & 32.31 & 22.53 & 21.80 & 28.94 & 24.55 & 28.57 & 34.71 & 22.08& 25.41& 31.48\\
Trigram (T) & 18.63& 19.92 & 25.81 & 18.60 & 19.90 & 25.29 & 18.60 & 19.95 & 25.73 & 18.60& 19.93& 25.29\\
U+B & 26.25& 32.80 & 35.34 & 27.18 & 32.27 & 36.68 & 28.82 & 31.27 & 35.14 & 23.96& 33.26& 33.99\\
B+T & 22.13& 23.32 & 31.13 & 23.11 & 22.54 & 28.61 & 24.53 & 26.95 & 31.51 & 21.49& 25.47& 32.85\\
U+B+T & 27.34& 34.87 & 40.00 & 27.53 & 31.86 & 39.93 & 28.49 & 35.90 & 41.30 & 26.32& 34.09& 36.95\\
Char 2-gram (C2) & 29.11& 36.27 & 36.33 & 28.60 & 35.76 & 39.51 & 31.72 & 34.61 & 36.39 & 27.68& 35.89& 37.45\\
Char 3-gram (C3) & 26.30& 34.23 & 37.18 & 23.97 & 34.78 & 39.44 & 28.21 & 35.08 & 36.09 & 24.72& 31.91& 38.26\\
Char 4-gram (C4) & 23.98& 30.84 & 37.45 & 25.19 & 30.76 & 37.74 & 25.29 & 30.39 & 35.06 & 21.67& 30.21& 36.34\\
Char 5-gram (C5) & 22.35& 27.33 & 32.36 & 21.96 & 26.99 & 35.34 & 24.42 & 26.87 & 34.92 & 22.06& 27.32& 32.72\\
C2+C3 & 26.34& 35.26 & 36.14 & 24.08 & 34.40 & 38.45 & 29.33 & 34.66 & 36.12 & 23.63& 34.87& 39.02\\
C3+C4 & 24.33& 35.72 & 38.32 & 24.46 & 34.31 & 39.15 & 26.12 & 32.94 & 36.67 & 22.93& 32.55& 38.36\\
C4+C5 & 23.27& 29.27 & 35.93 & 23.80 & 28.35 & 37.57 & 23.94 & 28.67 & 35.41 & 21.85& 29.44& 33.56\\
C2+C3+C4 & 25.89& 34.66 & 37.53 & 23.90 & 34.45 & 36.92 & 27.06 & 33.09 & 36.74 & 23.35& 32.94& 38.26\\
C3+C4+C5 & 23.11& 33.46 & 38.23 & 23.79 & 31.99 & 38.07 & 25.20 & 32.47 & 34.79 & 22.19& 32.25& 36.66\\
C2+C3+C4+C5 & 24.07& 32.66 & 37.99 & 24.27 & 32.66 & 36.95 & 26.79 & 32.02 & 35.66 & 23.00& 33.14& 37.09\\
U+B+C3+C4+C5 & 27.33& 32.53 & 38.20 & 26.14 & 32.66 & 40.25 & 29.10 & 32.03 & 40.49 & 27.30& 32.23& 38.44\\
U+B+C2+C3+C4+C5 & 27.26& 32.86 & 38.07 & 27.37 & 32.86 & 40.83 & 30.16 & 32.06 & 40.17 & 26.89& 34.81& 39.63\\
U+B+T+C2+C3+C4+C5 & 25.84& 32.68 & 38.79 & 26.33 & 33.77 & 41.13 & 29.04 & 30.47 & 39.97 & 25.78& 32.74& 40.68\\
Embeddings (E) & 31.38& 32.48 & 35.74 & 30.51 & 33.94 & 39.61 & 32.06 & 31.47 & 36.20 & 29.59& 34.18& 39.47\\
U+B+C2+C3+C4+C5+E & 26.88& 35.78 & 41.56 & 27.05 & 35.33 & 43.40 & 29.75 & 34.13 & 42.16 & 26.78& 39.02& 41.47\\
U+B+T+C2+C3+C4+C5+E & 26.98& 36.18 & 42.32 & 25.92 & 35.76 & 43.61 & 29.08 & 33.37 & 42.48 & 26.18& 36.75& 43.01\\ \hline
\end{tabular}%
}
\caption{F1 scores of ABSA Restaurant for Each method}
\label{ABSA_Restaurant_all}
\end{table}

\begin{table}
\centering
\tiny
\begin{tabular}{lcccccc}
\hline
\multirow{2}{*}{\textbf{Model}} & \multicolumn{3}{c}{\textbf{Normal, T}} & \multicolumn{3}{c}{\textbf{Augmented, T'}} \\ \cline{2-7} 
 & 15\% & 50\% & 100\% & 15\% & 50\% & 100\% \\ \hline
BanglaBERT          & 40.34    & 51.84    & 58.30  & 48.54           & 57.66           & 60.88  \\
Unigram (U)       & 39.26    & 45.46    & 47.60  & 39.90 & 46.11 & 49.34  \\
Bigram  (B)      & 32.57    & 38.86    & 41.37  & 33.16           & 39.52           & 44.29 \\ 
Trigram (T)       & 23.34    & 28.88    & 33.67  & 23.60           & 30.30           & 35.58 \\ 
U+B            & 38.52    & 43.20    & 46.71  & 39.71           & 45.48           & 49.74 \\ 
B+T            & 32.29    & 39.36    & 42.92  & 33.54           & 38.94           & 43.82 \\ 
U+B+T          & 39.42    & 45.12    & 47.95  & 40.21           & 46.23           & 50.98 \\
Char 2-gram (C2)    & 38.52    & 44.91    & 46.50  & 39.54           & 45.76           & 48.09 \\ 
Char 3-gram (C3)   & 40.34    & 47.20    & 48.09  & 40.83           & 47.16           & 50.50 \\ 
Char 4-gram (C4)  & 38.92    & 46.28    & 48.69   & 39.70       & 46.56       & 50.82\\ 
Char 5-gram (C5)   & 37.64    & 44.54    & 47.75   & 38.59       & 45.66       & 50.04\\ 
C2+C3          & 39.96    & 47.21    & 48.47   & 40.57       & 47.56       & 50.70\\ 
C3+C4          & 39.37    & 46.88    & 49.01   & 40.15       & 47.63       & 51.33\\ 
C4+C5          & 39.07    & 45.68    & 48.20   & 39.42       & 46.53       & 50.69\\ 
C2+C3+C4       & 39.58    & 47.00    & 48.64   & 40.20       & 47.66       & 51.33\\ 
C2+C3+C4+C5    & 39.42    & 46.69    & 48.81   & 39.60       & 47.47       & 51.44\\ 
U+B+C3+C4+C5   & 39.64    & 46.27    & 49.18   & 39.90       & 47.61       & 51.42\\
U+B+T+C2+C3+C4+C5 & 39.50 & 46.51 & 49.78      & 40.17       & 46.94       & 51.55\\
Embeddings (E)     & 42.59    & 45.08    & 47.53   & 40.44 & 47.28 & 51.57\\
U+B+C2+C3+C4+C5+E    & 41.01 & 48.19 & 51.82      & 41.65 & 48.69 & 52.78\\
U+B+T+C2+C3+C4+C5+E & 41.06 & 48.29 & 52.07 & 41.48 & 48.79 & 53.06\\ \hline
\end{tabular}
\caption{Comparison of F1 scores across all five datasets between Normal, T, and BDA augmented T' datasets from BDA}
\label{compare_normal_bda_all}
\end{table}

% \bibliographystyle{elsarticle-harv} 
% \bibliography{main}


\end{document}